\documentclass[sn-mathphys-num]{sn-jnl}% Math and Physical Sciences Numbered Reference Style 
\usepackage{algorithm}
\usepackage{algorithmic}
\usepackage{graphicx}%
\usepackage{multirow}%
\usepackage{amsmath,amssymb,amsfonts}%
\usepackage{amsthm}%
\usepackage[title]{appendix}%
\usepackage{xcolor}%
\usepackage{textcomp}%
\usepackage{manyfoot}%
\usepackage{booktabs}%
\usepackage{algorithm}%
\usepackage{listings}%

\usepackage{lmodern}

\usepackage{multirow}
\usepackage{booktabs}
\usepackage[misc]{ifsym}
\usepackage{float}
\usepackage{subfigure}
\usepackage{threeparttable}

\usepackage{indentfirst}
\usepackage{balance}
\usepackage{tikz}

\usepackage{algorithm}
\usepackage{array}
\usepackage{textcomp}
\usepackage{stfloats}

\usepackage{caption} 

\usepackage{url}
\usepackage{verbatim}

\usepackage{colortbl}
\usepackage{diagbox}
\definecolor{mygray}{gray}{.7}

\def\eg{{\em e.g.}}
\def\etal{{\em et al.}}

\graphicspath{{./fig/}}

\newcommand{\figref}[1]{Fig. \ref{#1}}
\newcommand{\tabref}[1]{Tab. \ref{#1}}

\newcommand{\myPara}[1]{\vspace{.05in}\noindent\textbf{#1}}

\newcommand{\bl}[1]{\textbf{#1}}

\usepackage{graphicx}
\usepackage{float}
\usepackage{xcolor, xpatch, hyperref}

\theoremstyle{plain}%
\makeatletter
\def\changeBibColor#1{%
    \in@{#1}{iscen2020eccv,dong2022cvpr,dong2024tpami} 
    \ifin@\color{red}\else\normalcolor\fi
}
\xpatchcmd{\@bibitem}{\item}{\changeBibColor{#1}\item}{}{}
\makeatother

\begin{document}

\title{Few-shot Class-Incremental Learning via Generative Co-Memory Regularization}


\author[1,2]{\sur{Kexin Bao}}\email{baokexin@iie.ac.cn}

\author[1]{\sur{Yong Li}}\email{liyong@iie.ac.cn}

\author[3]{\sur{Dan Zeng}}\email{dzeng@shu.edu.cn}

\author*[1]{\sur{Shiming Ge}}\email{geshiming@iie.ac.cn}

\affil[1]{\orgdiv{Institute of Information Engineering}, \orgname{Chinese Academy of Sciences}, \orgaddress{\city{Beijing}, \postcode{100085}, \country{China}}}

\affil[2]{\orgdiv{School of Cyber Security}, \orgname{University of Chinese Academy of Sciences}, \orgaddress{\city{Beijing}, \postcode{100049}, \country{China}}}

\affil[3]{\orgdiv{Department of Communication Engineering}, \orgaddress{\city{Shanghai}, \postcode{200040}, \country{China}}}


\abstract{Few-shot class-incremental learning (FSCIL) aims to incrementally learn models from a small amount of novel data, which requires strong representation and adaptation ability of models learned under few-example supervision to avoid catastrophic forgetting on old classes and overfitting to novel classes. This work proposes a generative co-memory regularization approach to facilitate FSCIL. In the approach, the base learning leverages generative domain adaptation finetuning to finetune a pretrained generative encoder on a few examples of base classes by jointly incorporating a masked autoencoder (MAE) decoder for feature reconstruction and a fully-connected classifier for feature classification, which enables the model to efficiently capture general and adaptable representations. Using the finetuned encoder and learned classifier, we construct two class-wise memories: representation memory for storing the mean features for each class, and weight memory for storing the classifier weights. After that, the memory-regularized incremental learning is performed to train the classifier dynamically on the examples of few-shot classes in each incremental session by simultaneously optimizing feature classification and co-memory regularization. The memories are updated in a class-incremental manner and they collaboratively regularize the incremental learning. In this way, the learned models improve recognition accuracy, while mitigating catastrophic forgetting over old classes and overfitting to novel classes. Extensive experiments on popular benchmarks clearly demonstrate that our approach outperforms the state-of-the-arts.}

\keywords{Few-shot class-incremental learning, generative representation, domain adaptation, object classification}



\maketitle

\section{Introduction}\label{sec:intro}

In many practical computer vision applications, \eg, visual object tracking~\cite{ross2008ijcv}, face recognition in video surveillance~\cite{yang2023tmm}, clinical data analysis~\cite{kiyasseh2021naturecomm} and remote sensing image classification~\cite{wang2024tgrs}, we often need to learn models from continuous information flow by incrementally incorporating new example knowledge. Typically, the model needs to be learned on a stream of few examples with incremental classes, which refers to few-shot class-incremental learning (FSCIL)~\cite{tao2020cvpr}. The FSCIL task aims to enable incremental learning with a small amount of data, which minimizes the need for additional labeled examples. However, incremental sessions often dilute or even discard previously learned knowledge while small examples of novel classes are easily remembered by the model, leading to catastrophic forgetting over old classes and overfitting to novel classes~\cite{tao2020cvpr}. 
Many approaches have been proposed to address that.

An intuitive solution is finetuning~\cite{sung2022cvpr,agarwal2022mm} by further training model with a small amount of data for adapting to the new task with some skill to deal with too few examples~\cite{chen2021iclr,chi2022cvpr}. Qi \etal~\cite{qi2018cvpr} applied class imprinting to variably alleviate catastrophic forgetting and flexibly separated distance between old and novel classes. Prototypical similarity methods~\cite{snell2017nips,yang2023tmm} remember new and old knowledge more stably. Such methods use the distance between the support set and the query set to reduce loss quickly and improve accuracy \cite{zhu2021cvpr}, which mitigate catastrophic forgetting by preserving features and reducing overfitting by fixing feature centers. These methods only focus on fast adaptation to novel classes or stable memories, which are easy to confuse old and novel classes due to overlooking the relationship between old and novel classes. Thus, when the model is adjusted, it needs a strong memory for old and new types of knowledge and keeps a sufficient distance between old and novel class examples.

\begin{figure}[t]
\centering
\includegraphics[width=0.70\linewidth]{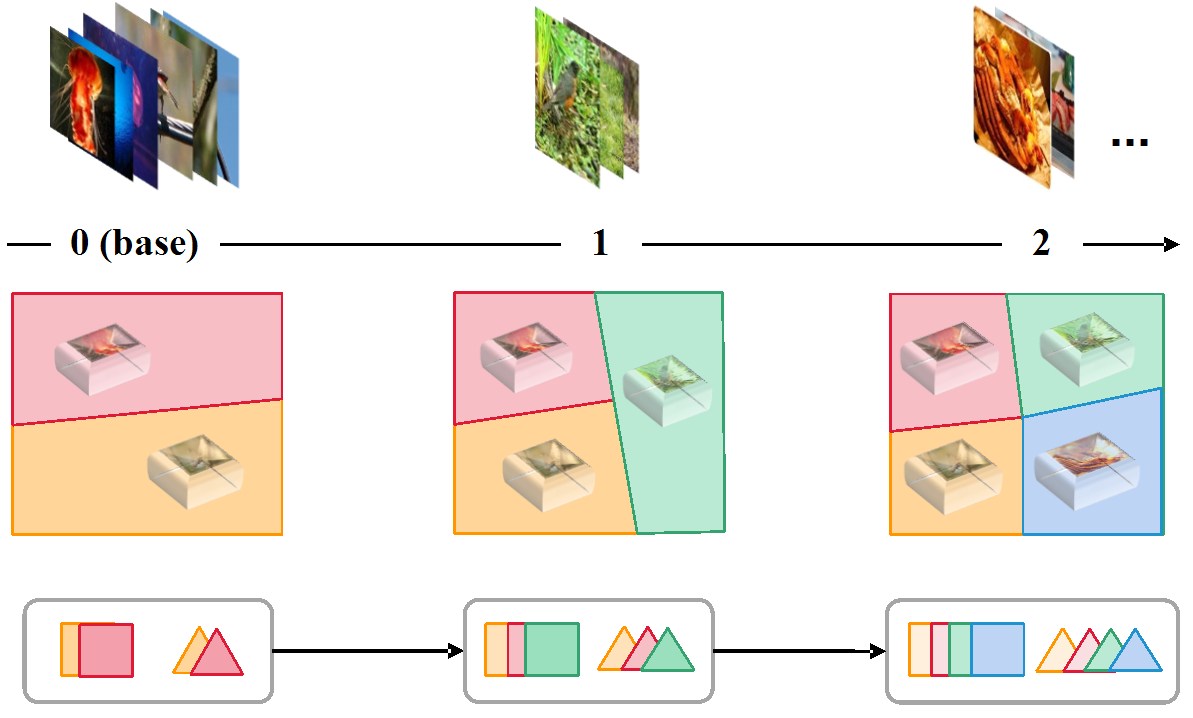}
\caption{We apply co-memory regularization in model learning, which facilitates discriminative classification while mitigating catastrophic forgetting and overfitting.}
\label{fig:motivation}
\end{figure}

Beyond catastrophic forgetting and overfitting, distribution discrepancy often exists in different datasets or the data with different classes, which consequently degrades the model in incremental learning tasks. To address that, learning a good base model is the key to adequately extract domain knowledge from a few examples to improve its representation adaptation. It ensures that datasets aggregate within the domain rather than adjust globally, which reduces the burden of training in incremental sessions. Generally, adequate domain knowledge always requires massive examples~\cite{wang2021ijcai}, which is not suitable for FSCIL. Domain adaptation provides an alternative to mine example knowledge and quickly adapt to the new data distribution, which is beneficial to knowledge extraction in both base and incremental learning.
 
In this paper, we propose a generative co-memory regularization approach to facilitate FSCIL, as shown in \figref{fig:motivation}. In base session, generative domain adaptation finetuning step finetunes a pretrained vision Transformer (ViT)~\cite{dosovitski2021iclr} encoder by jointly incorporating a masked autoencoder (MAE)~\cite{he2022cvpr} decoder for self-supervised feature reconstruction and a two-layer classifier for supervised feature classification. By combining two tasks, the finetuned encoder can learn general representations and realize effective adaptation. Then, the encoder is frozen to construct two cooperative memories: a representation memory for storing the mean feature of the examples in each class, and a weight memory for storing the classifier weights before the final layer. After that, memory-regularized incremental learning step incrementally updates the two memories and applies them to regularize the classifier learning on incremental datasets by: 1)~initializing classifier weights with weight memory and then optimizing the classification task, and 2) projecting representation memory and optimizing the regularization task. In this way, our approach incorporates domain knowledge and memory, leading to strong representation and generalization. 

Our main contributions are three folds. First, we propose a generative co-memory regularization approach to facilitate FSCIL by using generative features to construct representation memory and weight memory, which effectively mitigates catastrophic forgetting and overfitting. Second, we propose unsupervised domain adaptation finetuning by jointly optimizing self-supervised masked feature reconstruction and feature classification. Finally, we conduct extensive experiments on three benchmarks to show that our proposed approach outperforms the state-of-the-art methods.

\section{Related Works}

\myPara{Regularization-based FSCIL.} It aims to minimize the impact of the weights while training on new tasks and mitigate catastrophic forgetting. Recent studies solve the problem by penalizing parameter updates or knowledge distillation~\cite{wang2025tpami}. Some approaches via penalizing parameter updates repeat the consolidation of past knowledge by constraint parameters to strengthen the learned knowledge via memory-aware synapses~\cite{aljundi2018eccv}, slowing down the learning of important weights~\cite{kirkpatrick2017pnas}. These methods often remember old knowledge well but get poor at fitting new tasks when facing novel classes. Some methods penalize unimportant parameter updates to learn new tasks via a quadratic penalty in continual learning~\cite{lee2020cvpr,guo2023iccvw} or impose a small penalty on unimportant parameter updates~\cite{zenke2017icml}. These methods fit novel classes well but bring about faster forgetting of old classes. Some methods fit new tasks while retaining old knowledge by devising two regularization terms~\cite{ahn2019nips}, bi-level optimization~\cite{chi2022cvpr}, subspace regularization~\cite{akyurek2022iclr} or freezing certain crucial parameters~\cite{kim2023iclr}. MASIL~\cite{khandelwal2023tagml} use the idea of concept factorization explaining the collapsed features for base session classes in terms of concept basis and use these to induce classifier simplex for a few-shot classes. Zhao~\etal~\cite{zhao2023icme} use an imprinting-based distillation module for effectively regularizing the adaption process and a mathematically provable routing strategy for further improved results. Besides, another line of works~\cite{cui2023tmm} tends to fit new tasks and retain old knowledge with {knowledge distillation}. LDC~\cite{liu2023tpami} is built upon a parameterized calibration unit (PCU), which initializes biased distributions for all classes based on classifier vectors (memory-free) and a single covariance matrix. Lin~\etal~\cite{lin2023cvpr} incorporated information into the distillation process to enhance the memory in incremental sessions. Dhar~\etal~\cite{dhar2019cvpr} proposed attention distillation loss to preserve information about the base classes as novel classes are added. Recently, Dong \etal~\cite{dong2022cvpr,dong2024tpami} proposed a novel federated class-incremental learning framework that can be considered as a generalized regularization-based method. The method also penalizes the changes to important parameters but is different from our approach in its usage of global-local model for forgetting compensation rather than memory. In general, the existing regularization methods focus on parameter constraints but neglect feature generalization and memorizing old classes. 

\myPara{Memory-based FSCIL.}
It uses a small buffer of previous data replayed alongside new data~\cite{ji2023tip,chen2021iclr}. Some approaches add examples from old classes for subsequent training by selecting representative examples~\cite{rebuffi2017cvpr} or adaptive examples~\cite{lin2023tcsvt}. 
While these methods easily alleviate forgetting, they are also prone to expose privacy. The generative methods are effective in discouraging forgetting and protecting privacy by generating synthetic training examples~\cite{wu2018nips} or pseudo patterns~\cite{ostapenko2019cvpr}. The generated examples offset the lack of training examples, but there are still deviations between generated and real examples. DMP~\cite{tai2023tgrs} uses a prototypes-distillation (PD) network to learn to distill the features and prototypes into a lower dimensional in which ineffective features are eliminated. Some researchers improve architectures to promote memorization. Tao~\etal~\cite{tao2020cvpr} used a Neural Gas network to learn the topology of feature manifold. LPILC~\cite{bai2022tcyb} used a linear programming model to modify the classifier weights continuously. 
LIMIT~\cite{zhou2022tpami} built a generalizable feature space via a Transformer-based calibration module. Yang~\etal~\cite{yang2023iclr} proposed a neural collapse-inspired framework to achieve feature-classifier alignment in an incremental fashion, which requires prior knowledge of total class number. Iscen \etal~\cite{iscen2020eccv} regularized feature representations to retain old knowledge by only storing representations in memory. These methods have a strong generalization and fast updating, and learn new knowledge quickly but also may loss of old knowledge to a certain extent.
Besides, some methods calculate vectors from real features via retaining important knowledge~\cite{triki2017iccv}, introducing virtual prototypes~\cite{zhou2022cvpr} and compressing knowledge into a small number of quantized reference vectors~\cite{chen2021iclr}, thus can retain knowledge quickly with a small amount of memory. Inspired by this, we choose calculated features assisted by regularization to preserve old knowledge and quickly fit new data through the fully connected layer, which offers a practical advantage without prior knowledge requirement of total class number. 

\myPara{Unsupervised domain adaptation.} The task of FSCIL often suffers from domain-shift due to novel classes, thus unsupervised domain adaptation in representations can be used to facilitate FSCIL~\cite{kundu2020eccv}. Its objective is to adjust the domain differences so that the trained model has discriminative ability in new domain. Many FSCIL approaches~\cite{tao2020cvpr,zhou2022tpami,peng2022eccv,zhou2022cvpr} used ResNet as backbone and carefully designed loss functions to learn robust representations from scratch without pretraining in base session to achieve domain adaptation ability. Zhang \etal~\cite{zhang2022tcsvt} achieved domain adaptation by minimizing domain differences by adversarial learning, which speeds up the process of domain adaptation. Luo \etal~\cite{luo2023nn} fixed the parameters of the feature extractor and maximizes style discrepancy to update the classifier, which helps detect more hard examples. To improve few-shot learning and class-incremental learning tasks, recent several works \cite{zhai2023iccv,zhou2024cvpr,clap2024cvpr,park2024cvpr,wang2024ijcv,tang2024eccv} have been proposed to use vision transformer or pretraining method to improve FSCIL tasks, \eg, by using masked autoencoders \cite{zhai2023iccv}, pretrained vision transformer \cite{zhou2024cvpr} and large vision-language models \cite{clap2024cvpr,park2024cvpr}. Zhou \etal \cite{zhou2024ijcv} studied the class-incremental learning with pretrained models and stated the importance of generalizability and adaptivity. Typically, when a FSCIL task has limited base examples or uses a large network for backbone encoder, pretraining on a large dataset (\eg, ImageNet~\cite{deng2009cvpr}) is a standard practice for domain adaptation. For example, NC-FSCIL~\cite{yang2023iclr} and OrCo~\cite{ahmed2024cvpr} apply pretraining on ImageNet to perform FSCIL task on CUB200 with less than 6000 base examples, while PriViLege~\cite{park2024cvpr} and CPE-CLIP~\cite{alessandro2023iccvw} use ViT as backbone and pretrains it on ImageNet. In general, the key of domain adaptation in a FSCIL task is learning more robust and generalizable representations for general objects, which inspires us to learn generative representations with a vision transformer and masked modeling in the encoder in an self-supervised manner rather than supervised pretraining in~\cite{tao2020cvpr,zhang2021cvpr,yang2023iclr,ahmed2024cvpr}.
  
\section{Our Approach}
\subsection{Problem Formulation}\label{Problem}
We follow vanilla few-shot class-incremental learning (FSCIL)~\cite{tao2020cvpr}, where there are a stream of labeled training sets $\mathcal{D}=\{D^{(t)}\}_{t=0}^{T}$ with $D^{(t)}=\{(\bl{x}^{(t)}_j,y^{(t)}_j)\}_{j=1}^{|D^{(t)}|}$. Here, $\bl{x}^{(t)}_j$ is an image and $y^{(t)}_j\in L^{(t)}$ denotes its class label where $L^{(t)}$ is the class set of $t$-th training set. Any two class sets are disjoint, meaning $L^{(i)}\cap L^{(j)}=\emptyset$ for $\forall i, j$. 
In our setting, $D^{(0)}$ is base training set and $D^{(t)} (t>0)$ is few-shot training set of novel classes in the $t$-th session. The task of FSCIL is learning a model $\phi(\bl{x};\bl{w})$ with parameters $\bl{w}$ that incrementally trains on $\mathcal{D}$ where only $D^{(t)}$ is available in the $t$-th training session. After the $t$-th training session on $D^{(t)}$, the model $\phi(\bl{x};\bl{w})$ is tested to recognize the classes in $\cup_{i=0}^{t} L^{(i)}$. In general, for $D^{(t)}, t>0$, the setting of learning task is described by \emph{$N$-way $K$-shot} FSCIL where $N$ is the class number and $K$ represents the number of training examples per class.

\begin{figure*}[t]
\centering
\includegraphics[width=1.0\linewidth]{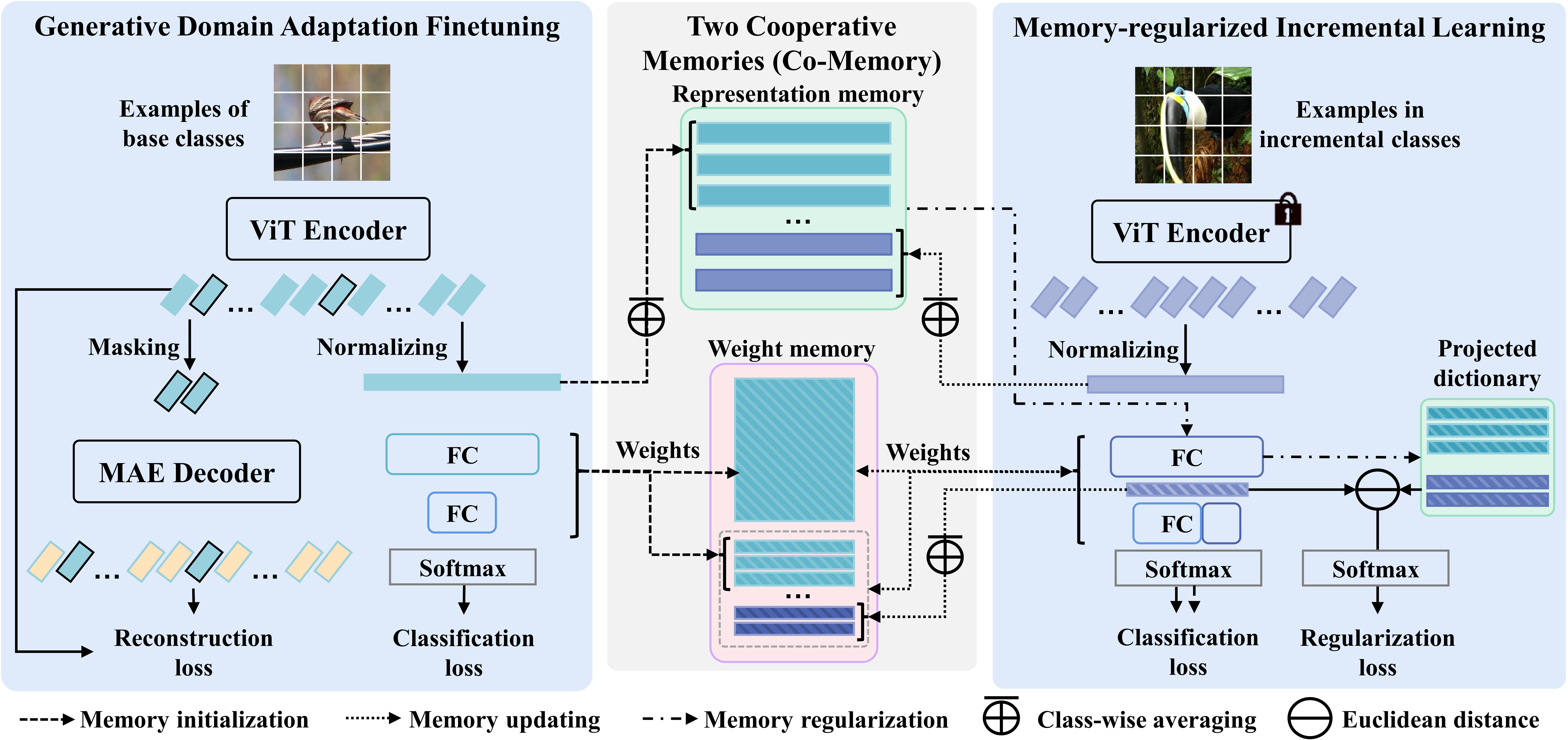}
\vspace{-10pt}
\caption{The framework of our approach consists of two steps. The generative domain adaptation finetuning step finetunes a pretrained ViT encoder on the base classes by jointly incorporating a MAE decoder for self-supervised masked feature reconstruction and a two-layer classifier for supervised feature classification. Then, the ViT encoder is frozen and used to construct two cooperative memories with image examples: representation memory which stores the class mean features over each class, and weight memory which stores the classifier weights of two fully-connected layers before the final softmax layer. Finally, the memory-regularized incremental learning step updates two memories and uses them to collaboratively regularize the incremental learning in each session explicitly and implicitly, respectively.}
\vspace{-10pt}
\label{fig:method}
\end{figure*}

Typically, $D^{(0)}$ is not enough to directly train a deep learning model. Thus, we need to train on $D^{(0)}$ to get a high-accurate model that can further incrementally learn on $D^{(t)} (t>0)$ without catastrophic forgetting over old classes $\cup_{i=0}^{t-1} L^{(i)}$ and overfitting to novel classes $L^{(t)}$. To this end, we use the pretrained model and finetune it on $D^{(0)}$ for domain adaptation and perform few-shot incremental learning on $D^{(t)} (t>0) $ assisted by external memories. In our setting, the model $\phi(\bl{x};\bl{w})$ consists of an encoder $\phi_e(\bl{x};\bl{w}_e)$ with parameters $\bl{w}_e$ and an incremental classifier $\phi_c(\bl{f};\bl{w}_c)$ with parameters $\bl{w}_c$, and $\bl{w}=\bl{w}_e\cup\bl{w}_c$. Here $\bl{f}=\phi_e(\bl{x};\bl{w}_e)$ is the encoded features of $\bl{x}$. By introducing an external memory $\bl{M}$, the model learning can be formulated as: 
\begin{equation}
\begin{aligned}
\{\bl{w}_e,\bl{w}_c^{(0)}\}&=\arg\min_{\{\bl{w}_e,\bl{w}_c^{(0)}\}}\mathbb{E}_b(D^{(0)};\phi_e,\bl{w}_e,\phi_c,\bl{w}_c^{(0)}), \\
\{\bl{w}_c^{(t)},\bl{M}\}&=\arg\min_{\bl{w}_c^{(t)}}\mathbb{E}_i(D^{(t)};\phi_e,\bl{w}_e,\phi_c,\bl{w}_c^{(t)};\bl{M}).
\end{aligned}
\label{eq:problem}
\end{equation}
Obviously, it consists of learning on base classes and incremental learning on a stream of novel classes, which both suffer from inadequate examples. Thus, the model needs to adequately extract knowledge from few examples in $D^{(0)}$ in the base learning, while it needs regularization to avoid overfitting to $D^{(t)}$ in the incremental learning. To this end, our approach applies two major steps to solve base learning and incremental learning via generative domain adaptation finetuning and memory-regularized incremental learning, respectively, while the knowledge is retained and transferred between these two steps by using two external memories, as shown in Fig. \ref{fig:method}. 

\subsection{Generative Domain Adaptation Finetuning }\label{finetuning}
In base session, generative domain adaptation finetuning takes the encoder of a pretrained visual transformer (ViT)~\cite{dosovitski2021iclr} as the initial backbone $\phi_e(\bl{x};\bl{w}_e)$ and finetunes its parameters $\bl{w}_e$ on the training dataset of base classes $D^{(0)}$. The finetuning should address effective domain adaptation over a few examples since the size of dataset $|D^{(0)}|$ is often not sufficient to directly train a powerful ViT. To this end, our finetuning is proposed to make full use of $D^{(0)}$ in a semi-supervised manner, which is performed by incorporating the decoder of masked autoencoders (MAE) $\phi_d(\bl{f};\bl{w}_d)$ with parameters $\bl{w}_d$ for self-supervised feature reconstruction and a classifier consisting of two fully-connected layers and a softmax layer for supervised feature classification. The classifier uses a single fully-connected layer to integrate the encoder characteristics, which alleviates the effect of unimportant features, while using more layers or complex structures will make feature fitting with few examples difficult. It has been shown~\cite{zhang2018pcm} that when the target domain has small data or the difference between source and target domains is large, fully-connected layers play an important role in achieving high target accuracy by model finetuning. 

During training, each example is patchified and fed into the ViT encoder to generate a group of features. For feature reconstruction, $75\%$ features are randomly masked out and the remaining features are fed into MAE decoder for self-supervised reconstruction, where the reconstruction loss is measured by the difference between the reconstructed and input feature groups. For feature classification, the feature group is first normalized and integrated into a feature vector and then fed into {a $|L^{(0)}|$-way fully-connected layer} and a softmax layer to require its classification prediction. Thus, we update all model parameters $\{\bl{w}_e,\bl{w}_d,\bl{w}_c^{(0)}\}$ by minimizing the total base loss $\mathbb{E}_b$:
\begin{equation}
    \label{eq:finetuning}
    \begin{aligned}
\mathbb{E}_b(D^{(0)};\bl{w}_e, \bl{w}_d, \bl{w}_c^{(0)})&= \alpha\sum_{j=1}^{|D^{(0)}|}{\frac{|\phi_d(\hat{\bl{f}}_{j}^{(0)};\bl{w}_d)-\phi_e(\bl{x}_j^{(0)};\bl{w}_e^{(0)})|^2}{|D^{(0)}|}}\\
    &+(1-\alpha)\sum_{j=1}^{|D^{(0)}|}\ell_{CE}(\phi_c(\overline{\bl{f}}_{j}^{(0)};\bl{w}_c^{(0)}),y_j^{(0)}),
    \end{aligned}
\end{equation}
where $\hat{\bl{f}}_{j}^{(0)}$ and $\overline{\bl{f}}_{j}^{(0)}$ are the masked and normalized features of $\bl{f}_j^{(0)}=\phi_e(\bl{x}_j^{(0)};\bl{w}_e^{(0)})$, respectively. $\ell_{CE}(\cdot)$ measures the cross-entropy loss.  $\alpha = c\sqrt{(1/e)^{E}}$ is controlled by the training epochs $E$, where $c$ is the ratio of classification and reconstruction loss. The formula first uses a large proportion of reconstruction for quickly adapting to domain knowledge, and then reduces the reconstruction influence.    

\subsection{Memory-regularized Incremental Learning}\label{learning}
After finetuning, the encoder is frozen to enable effective few-shot learning~\cite{tsimpoukelli2021nips,qiang2023ijcv,wu2024ijcv} under large parameters and few examples. The main task of memory-regularized incremental learning is incrementally training the classifier $\bl{w}_c$ on $D^{(t)}$ and facilitating model performance to recognize the classes in $\cup_{i=0}^{t} L^{(i)}$. In incremental sessions, classifier training under few examples of novel classes inevitably causes misalignment between the features and classifier weights of old classes. Thus we apply two cooperative memories (co-memory) $\bl{M}=\{\bl{M}_e,\bl{M}_w\}$, including representation memory $\bl{M}_e$ and weight memory $\bl{M}_w$ to regularize the incremental training. 

\myPara{Memory initialization and updating.}~The memories are initialized with the ViT encoder and classifier finetuned on $D^{(0)}$ after base session, and are updated on $D^{(t)}$ after the $t$-th incremental session. For representation memory, all the examples of each class in $D^{(0)}$ are fed into the ViT encoder and the normalization layer, and then the features are averaged into a $768$-dimensional vector. Representation memory stores all the class mean feature vectors, leading to an initial $|L^{(0)}|$-by-$768$ matrix $\bl{M}_e^{(0)}$, and then are updated by incrementally adding new class mean feature vectors to get a $|\cup_{i=0}^{t} L^{(i)}|$-by-$768$ matrix $\bl{M}_e^{(t)}$. For weight memory, $\bl{M}_w^{(0)}=\{\bl{M}_f^{(0)},\bl{M}_p^{(0)}\}$ initially stores the classifier weights $\bl{M}_f^{(0)}=\bl{w}_c^{(0)}$ and a $|L^{(0)}|$-by-$256$ matrix $\bl{M}_p^{(0)}=\phi_c(\bl{M}_e^{(0)};\hat{\bl{w}}_c^{(0)})$ by projecting  $\bl{M}_e^{(0)}$ with the classifier weights before the last fully-connected layer $\hat{\bl{w}}_c^{(0)}\subset\bl{w}_c^{(0)}$. We then use the classifier weights before the last fully-connected layer and re-calculate projected features to update weight memory, leading to $\bl{M}_f^{(t)}$ and a $|\cup_{i=0}^{t} L^{(i)}|$-by-$256$ matrix $\bl{M}_p^{(t)}$. In this way, the memory footprint of the co-memory is fixed in base session and continuously increases in incremental sessions. However, there are no memory restrictions on general FSCIL settings when continuously learning multiple tasks, \eg, the memory only increases about $3$ MB even the total class number reaches $|\cup_{i=0}^{T} L^{(i)}|=1000$ in last session.

\myPara{Training with memory regularization.} With representation memory $\bl{M}_e^{(t)}$ and weight memory $\bl{M}_w^{(t)}$, we initialize the classifier parameters  $\bl{w}_c^{(t)}=\bl{M}_f^{(t)}$, and project $\bl{M}_e^{(t)}$ to construct the projected dictionary $\overline{\bl{M}}_e^{(t)}=\phi_c(\bl{M}_e^{(t)};\hat{\bl{w}}_c^{(t)})$ where $\hat{\bl{w}}_c^{(t)}\subset\bl{w}_c^{(t)}$. Then, we update the classifier parameters $\bl{w}_c^{(t)}$ by simultaneously optimizing feature regularization and feature classification to minimize the incremental loss $\mathbb{E}_i$:
\begin{equation}
   \label{eq:comemory}
   \begin{aligned}
\mathbb{E}_i(D^{(t)},\bl{M}_e^{(t)};\bl{w}_c^{(t)})&= \left[\beta\sum_{j=1}^{|D^{(t)}|}\ell_{CE}(\bl{d}_j^{(t)},y_j^{(t)})+(1-\beta){\sum_{k=1}^{|\bl{M}_e^{(t)}|}\ell_{CE}(\bl{m}_k^{(t)},k)}\right] \\
    &+(1-\beta)\sum_{j=1}^{|D^{(t)}|}\ell_{CE}(\phi_c(\overline{\bl{f}}_{j}^{(t)};\bl{w}_c^{(t)}),y_j^{(t)}),
    \end{aligned}
\end{equation}
where the first term is the regularization loss that contributes to representation regularization to mitigate catastrophic forgetting of old knowledge, the second term is the classification loss that enables the model to learn new knowledge, and the factor $\beta\in[0,1]$ is used to balance the effect of the two loss terms. $\phi_c(\overline{\bl{f}}_{j}^{(t)};\bl{w}_c^{(t)})$ is the classification output of the $j$-th incremental example. $\bl{m}_{k}^{(t)}=\phi_c(\bl{M}_e^{(t)}(k);\bl{w}_c^{(t)})$ is the classification output of the $k$-th feature vector in representation memory. $\bl{d}_j^{(t)}$ is a vector whose element is the Euclidean distance between the normalized feature vector of example and each feature vector in $\bl{M}_e^{(t)}$:
\begin{equation}
\label{eq:memory}
         \bl{d}_j^{(t)}(k)=|\overline{\bl{f}}_{j}^{(t)}-\overline{\bl{M}}_e^{(t)}(k)|^2.
\end{equation}
The regularization loss in Eq.~\eqref{eq:comemory} contains two parts. On one hand, the first part encourages separation between representations of novel examples from current session and memory representations, thereby improving class discriminability. On the other hand, the second part ensures the memory representations are correctly classified, thereby enabling the model to remember old class knowledge.

\begin{center}
\vspace{-5pt}
\begin{algorithm}[ht]
    \caption{Generative Co-Memory Regularization (GCMR)}
    \label{alg:algorithm}
    {\textbf{Input}: The training datasets $\{D^{(t)}\}_{t=0}^{T}$, the label sets $\{L^{(t)}\}_{t=0}^{T}$, the model $\phi=\{\phi_e,\phi_c\}$, the factor $\alpha$ and $\beta$, representation memory $\bl{M}_e$ and weight memory $\bl{M}_w$.}\\
    {\textbf{Output}: The model $\phi(\bl{x};\bl{w}^{(t)})$ and memory $\bl{M}^{(t)}=\{\bl{M}_e^{(t)},\bl{M}_w^{(t)}\}$.}\\
    \vspace{-10pt}
    {
    \begin{algorithmic}[1]
        % \STATE Let $t=0$;
        \STATE Initialize the backbone encoder with pretrained model.
        \FOR{$t=0$ to $T$} 
        %\STATE Input the dataset $\mc{D}^{(0)}$ and 
        \IF{$t=0$}
        \STATE Compute feature $\bl{f}_j^{(0)}=\phi_e(\bl{x}_j^{(0)},\bl{w}_e)$ for each example on base training set $D^{(0)}$, and then achieve $\hat{\bl{f}}_{j}^{(0)}$ with masking and $\overline{\bl{f}}_{j}^{(0)}$ with normalizing; 
        \STATE Compute outputs via the classifier $\phi_c(\overline{\bl{f}}_{j}^{(0)},\bl{w}_c^{(0)})$ and the decoder $\phi_d(\hat{\bl{f}}_{j}^{(0)};\bl{w}_d)$;
        \STATE Update model parameters $\bl{w}^{(0)}=\{\bl{w}_e,\bl{w}_d,\bl{w}_c^{(0)}\}$ by optimizing $\mathbb{E}_b$ in Eq.~\eqref{eq:finetuning}.
        \ELSE
        \STATE Freeze $\phi_e(\bl{x}_j,\bl{w}_e)$, and initialize $\phi_c(\overline{\bl{f}}_{j}^{(t)},\bl{w}_c^{(t)})$ by $\bl{w}_c^{(t)}=\bl{M}_f^{(t)}\subset\bl{M}_w^{(t)}$;
        \STATE Compute feature $\bl{f}_j^{(t)}=\phi_e(\bl{x}_j^{(t)},\bl{w}_e)$ for each example on incremental training set $D^{(t)}$, and then achieve $\overline{\bl{f}}_{j}^{(t)}$ with normalizing;
        \STATE Project representation memory $\bl{M}_e^{(t)}$ to get the projected dictionary $\overline{\bl{M}}_e^{(t)}$, and then compute vectors $\bl{d}_j^{(t)}$ according to Eq.~\eqref{eq:memory};
        \STATE Compute predictions $\phi_c(\overline{\bl{f}}_{j}^{(t)},\bl{w}_c^{(t)})$ and $\bl{m}_{k}^{(t)}=\phi_c(\bl{M}_e^{(t)}(k);\bl{w}_c^{(t)})$;
        \STATE Update the classifier parameters $\bl{w}_c^{(t)}$ by optimizing $\mathbb{E}_i$ in Eq.~\eqref{eq:comemory};
        \ENDIF 
        \STATE Compute the class mean features on $D^{(t)}$ and add them into $\bl{M}_e^{(t)}$;
        \STATE Compute $\bl{M}_p^{(t)}$ and combine with classifier weights $\bl{M}_f^{(t)}$ to get $\bl{M}_w^{(t)}=\{\bl{M}_f^{(t)},\bl{M}_p^{(t)}\}$.
        \ENDFOR
        % for loop
    \end{algorithmic}
    }   
\end{algorithm}
\vspace{-10pt}
\end{center}

\subsection{Discussion}
\myPara{Base learning.}~The ViT encoder serves as the foundation in both base and incremental learning due to its powerful representation ability~\cite{dosovitski2021iclr}. We finetune a ViT encoder by jointly self-supervised feature reconstruction and supervised feature classification, which effectively improves data utilization to reduce the dependence on large amounts of data~\cite{steiner2022tmlr} and force the model to learn discriminative semantic features rather than the superficial surrounding information~\cite{he2022cvpr}. Specifically, feature masking and reconstruction processes like MAE~\cite{he2022cvpr} and SMKD~\cite{lin2023cvpr} can improve the fitting efficiency, leading to a fast knowledge adaptation to the target domain.

\myPara{Incremental learning.}~Two external memories are used to assist incremental learning. Weight memory is applied to classifier initialization like~\cite{qi2018cvpr} for further finetuning, where class mean features and weights are symmetric in the fully-connected layer, making the mean with each novel class closer to the corresponding training weights. Intuitively, the averaging features are the templates for remembering the semantic embedding of low-shot examples in novel classes. Representation memory stores the class mean features that act as prototypes~\cite{snell2017nips,yang2023iclr} and can stably preserve important features and high feature alignment. 

\myPara{Generative co-memory regularization.}~Alg.~\ref{alg:algorithm} gives the algorithm details of our generative co-memory regularization approach. In base session, it integrates a ViT~\cite{dosovitski2021iclr} encoder as backbone and a MAE~\cite{he2022cvpr} decoder for self-supervised feature reconstruction to perform finetuning in the context of FSCIL. There are three major differences from conventional ViT and MAE: 1) its objective is finetuning the pretrained ViT encoder on base examples to achieve domain-adapted representations; 2) it performs masked modeling in feature level rather than pixel level of traditional MAE framework, which can refine features to achieve task-adapted representations; and 3)  it further applies a co-memory mechanism in incremental sessions to jointly constrain parameter update, consisting of a representation memory for preserving class prototypes and a weight memory for retaining classifier knowledge. Our generative co-memory regularization combines the strengths of memory-based and regularization-based continual learning methods. On one hand, representation memory explicitly regularizes the classifier by preserving past class prototypes through a frozen encoder. On the other hand, weight memory implicitly regularizes training by initializing the updated classifier with previously learned knowledge, thereby accelerating convergence and reducing overfitting. The co-memory effectively mitigates catastrophic forgetting and leads to improved performance in the FSCIL setting.  

\section{Experiments}
To verify the effectiveness of our approach (denoted as \textbf{GCMR}), we conduct the experiments on three datasets. We first check the effectiveness of our incremental learning method with co-memory regularization. Then, we compare our GCMR to the state-of-the-art approaches.

\subsection{Experimental Settings}\label{Experimental Detail}
\myPara{Datasets.} Three FSCIL benchmarking datasets are used. MiniImageNet \cite{deng2009cvpr} and CIRAR100~\cite{krizhevsky2019cifar} have 100 classes with each class containing 500 training images and 100 testing images. CUB200~\cite{Wah2011TheCB} is a dataset for fine-grained image classiﬁcation containing 11,788 images of 200 classes. We adopt AugMix~\cite{hendrycks2020iclr} as the augmentation scheme including random translation, rotation and contrast enhancement. We follow the standard experimental settings in FSCIL~\cite{tao2020cvpr,zhang2021cvpr}. For both MiniImageNet and CIRAR100, base session contains 60 classes and a \emph{$5$-way $5$-shot} setting is adopted for each session, which means the other 40 classes are divided into 8 incremental sessions. For CUB200, 100 classes are used in base session, and the other 100 classes are assigned as 10 incremental sessions with the setting of \emph{$10$-way $5$-shot}. The class and example assignments are random. 

\myPara{Networks.} In default, we use a ViT-B~\cite{dosovitski2021iclr} pretrained on ImageNet with self-supervised MAE~\cite{he2022cvpr} as the backbone encoder, and finetune it in base session for domain adaptation. To ensure a fair evaluation, we compare the ViT-B encoder with the encoders of three ResNet architectures \cite{he2016cvpr} (ResNet34, ResNet18 and ResNet12 with or without pretraining) where the feature reconstruction branch is deleted. After a $768d$ normalization layer, we use a $256d$ fully-connected layer with ReLU activations to decode the encoded features, along with a $|\cup_{i=0}^{t} L^{(i)}|$-way fully-connected layer and a softmax layer for the final prediction.  

\myPara{Implementation.} Similar to recent works \cite{park2024cvpr,zhou2024cvpr}, our ViT-B is pretrained on ImageNet~\cite{deng2009cvpr}. We use SGD optimizer with momentum and adopt a cosine annealing strategy for learning rate. We train all models with a batch size of $32$. On CIRAR100 and MiniImageNet, we train 20 to 100 epochs in base session, and 50 to 100 epochs in each incremental session, with an initial learning rate of $0.001$ for base session and $0.002$ for incremental sessions. On CUB200, we train $100$ epochs in base session, 100 to 200 in each incremental session, with an initial learning rate of $0.002$ for base session and $0.001$ for incremental sessions. The factor $c$ is $0.1$ for MiniImageNet, $0.3$ for CIRAR100 and $0.5$ for CUB200 that affects $\alpha$ in Eq.~\eqref{eq:finetuning}. In our experiments, we set $\beta$ to 0.8, 0.7 and 0.7 for MiniImage, CIFAR100 and CUB200, respectively. {All input images are rescaled into $224\times224$ for the encoders of ViT-B, ResNet34 and ResNet18}, and then grid-wisely pacified into $196$ patches with $16\times16$ for ViT-B encoder. For a fair comparison, similar to \cite{yang2023iclr}, we rescale the input images into $84\times84$ for MiniImageNet and $32\times32$ for CIRAR100, respectively.

\subsection{The Effectiveness of Co-Memory Regularization}\label{subsec:effect-dmr}

\begin{table*}[t]
\centering
\caption{Effectiveness of our co-memory regularization for incremental learning. All approaches use ResNet12 as the encoder without pretraining for a fair comparison.}
\resizebox{1.0\linewidth}{!}{
\begin{tabular}{ccccccccccccc}
\hline
\multirow{2}{*}{Dataset} & \multirow{2}{*}{Approach} &\multicolumn{9}{c}{Accuracy in each session ($\%$)  $\uparrow$} & Avg \\
     & & 0 & 1 & 2 & 3 & 4 & 5 & 6 & 7 & 8 & acc.\\
	\hline
\multirow{4}{*}{CIRAR100} 
        &{C-FSCIL~\cite{hersche2022cvpr} (CVPR'22)} & {77.47} & {72.40} & {67.47} & {63.25} & {59.84} & {56.95}& {54.42} & {52.47} & {50.47} & {61.64} \\
        &NC-FSCIL~\cite{yang2023iclr} (ICLR'23)  & 82.52 & 76.82 & 73.34 & 69.68 & 66.19 & 62.85 & 60.96 & 59.02 & 56.11 & 67.50 \\
        &{OrCo~\cite{ahmed2024cvpr} (CVPR'24)} &{80.08} &{68.16} &{66.99} &{60.97} &{59.78} &{58.60} &{57.04} &{55.13} &{52.19} &{62.11}\\
    & Our GCMR  & \textbf{82.61} & \textbf{76.92} & \textbf{73.35} & \textbf{69.83} & \textbf{66.58} & \textbf{63.17} & \textbf{61.08} & \textbf{59.10} & \textbf{56.24} & \textbf{67.65} \\
\hline
\multirow{3}{*}{MiniImageNet} 
        &{C-FSCIL~\cite{hersche2022cvpr} (CVPR'22)} & {76.40} & {71.14} & {66.46} & {63.29} & {60.42} & {57.46} & {54.78} & {53.11} & {51.41} & {61.61} \\
        &NC-FSCIL~\cite{yang2023iclr} (ICLR'23) & 84.02 & 76.80 & 72.00 & 67.83 & 66.35 & 64.04 & 61.46 & 59.54 & 58.31 & 67.82 \\
        & Our GCMR & \textbf{84.23} & \textbf{77.06} & \textbf{72.11} & \textbf{68.00} & \textbf{66.57} & \textbf{64.20} & \textbf{61.83} & \textbf{59.81} & \textbf{58.53} & \textbf{68.04} \\  
\hline
\end{tabular}
}
\label{tab:effect-dmr}
\vspace{-10pt}
\end{table*}

To evaluate the effectiveness of our incremental learning method, we conduct an experimental comparison with the recent competing approaches on CIFAR100 and MiniImageNet. To make a fair evaluation, we use the same ResNet12 structure in NC-FSCIL~\cite{yang2023iclr} as the encoder and initialize it randomly without pretraining, and take the input images with the same size. Here, ResNet12 is a ResNet variant~\cite{hersche2022cvpr} which contains four residual blocks with 64, 160, 320 and 640 filters, leading to more parameters than standard ResNet18. From the results reported in Tab. \ref{tab:effect-dmr}, compared to other competing approaches, we can see that our approach performs better in base and all incremental sessions. Our approach involves adding some constraints to the training process to take into account the correlations between outputs in previous and current sessions. The constraints allow the model to learn a more general and robust feature representation that can improve its generalization ability in incremental sessions. By applying the co-memory regularization, the performance is improved in both base session and all incremental sessions, which proves the effectiveness of improving the stability and generalization ability of our co-memory regularization.

\begin{table*}[t]
\caption{FSCIL performance on CIRAR100. ``Avg imp.'' calculates our improvement over other approaches. The approaches include class-incremental or few-shot learning approaches with FSCIL setting, and FSCIL approaches.}
\label{tab:cifar100}
\centering
\resizebox{\linewidth}{!}{
\begin{tabular}{lccccccccccc}
\hline
	\multirow{2}{*}{Approach} & \multicolumn{9}{c}{Accuracy in each session ($\%$)  $\uparrow$} & Avg & Avg \\
    &  0 & 1 & 2 & 3 & 4 & 5 & 6 & 7 & 8 & acc. & imp.\\
	\hline
 D-Cosine \cite{vinyals2016nips} (NeurIPS'16) & 74.55 & 67.43 & 63.63 & 59.55 & 56.11 & 53.80 & 51.68 & 49.67 & 47.68 & 58.23 & +11.63\\
	iCaRl \cite{rebuffi2017cvpr} (CVPR'17) & 64.10 & 53.28 & 41.69 & 34.13 & 27.93 & 25.06 & 20.41 & 15.48 & 13.73 & 32.87 & +36.9 \\
        NCM \cite{hou2019cvpr} (CVPR'19) & 64.10 & 53.05 & 43.96 & 36.97 & 31.61  & 26.73 & 21.23 & 16.78 & 13.54 & 34.22 & +35.55 \\
        \hline
        TOPIC \cite{tao2020cvpr} (CVPR'20) & 64.10 & 55.88 & 47.07 & 45.16 & 40.11 & 36.38 & 33.96 & 31.55 & 29.37 & 42.62 & +27.15 \\
        SPPR \cite{zhu2021cvpr} (CVPR'21) & 64.10 & 65.86 & 61.36 & 57.45 & 53.69 & 50.75 & 48.58 & 45.66 & 43.25 & 54.52 & +15.25 \\
        CEC \cite{zhang2021cvpr} (CVPR'21) & 73.07 & 68.88 & 65.26 & 61.19 & 58.09 & 55.57 & 53.22 & 51.34 & 49.14 & 59.53 & +10.24 \\    
        LIMIT \cite{zhou2022tpami} (TPAMI'22) & 73.81 & 72.09 & 67.87 & 63.89 & 60.70 & 57.77 & 55.67 & 53.52 & 51.23 & 61.84 & +7.93 \\
        MetaFSCIL \cite{chi2022cvpr} (CVPR'22) & 74.50 & 70.10 & 66.84 & 62.77 & 59.48 & 56.52 & 54.36 & 52.56 & 49.97 & 60.79 & +8.98 \\
        C-FSCIL \cite{hersche2022cvpr} (CVPR'22) & 77.47 & 72.40 & 67.47 & 63.25 & 59.84 & 56.95 & 54.42 & 52.47 & 50.47 & 61.64 & +8.33\\
        DF Replay \cite{liu2022eccv} (ECCV'22) & 74.40 & 70.20 & 66.54 & 62.51 & 59.71 & 56.58 & 54.52 & 52.39 & 50.14 & 60.78 & +8.99\\
        ALICE \cite{peng2022eccv} (ECCV'22) & 79.00 & 70.50 & 67.10 & 63.40 & 61.20 & 59.20 & 58.10 & 56.30 & 54.10 & 63.21 & +6.56\\
        DSN \cite{yang2023tpami} (TPAMI'23) & 73.00 & 68.83 & 64.82 & 62.64 & 59.36 & 56.96 & 54.04 & 51.57 & 50.00 & 60.14 & +9.63 \\
        MCNet~\cite{ji2023tip} (TIP'23)&73.30&69.34 &65.72&61.70&58.75 &56.44 &54.59 &53.01 &50.72 &60.40 &+9.37 \\
        TEEN \cite{wang2023nips} (NeurIPS'23) &74.92 &72.65 &68.74 &65.01 &62.01 &59.29 &57.90 &54.76 &52.64 &63.10 &+6.67 \\
        NC-FSCIL \cite{yang2023iclr} (ICLR'23) & 82.52 & 76.82 & 73.34 & 69.68 & 66.19 & 62.85 & 60.96 & 59.02 & 56.11 & 67.50 & +2.27\\
        GKEAL \cite{zhuang2023cvpr} (CVPR'23) &74.01 &70.45 &67.01 &63.08 &60.01 &57.30 &55.50 &53.39 &51.40 &61.35 & +8.42 \\
        CABD \cite{zhao2023cvpr}(CVPR'23)&79.45&75.38 &71.84 &67.95 &64.96 &61.95 &60.16 &57.67 &55.88 &66.14 &+3.63 \\
        OSHHG \cite{cui2024tmm} (TMM'24)&63.55&62.88 &61.05 &58.13 &55.68 &54.59 &52.93 &50.39 &49.48 &56.52 &+13.25 \\
        EHS \cite{deng2024wacv} (WACV'24)&71.27&67.40 &63.87 &60.40 &57.84 &55.09 &53.10 &51.45 &49.43 &58.87 &+10.90 \\
        OrCo~\cite{ahmed2024cvpr} (CVPR'24) & 80.08 &  68.16 & 66.99 &60.97 &59.78 &58.60 &57.04 &55.13 &52.19 &62.11& +6.45\\
	%\hline
	 \textbf{Our GCMR} & \textbf{85.52} & \textbf{77.97} & \textbf{74.50} & \textbf{71.19} & \textbf{68.69} & \textbf{66.07} & \textbf{63.59} & \textbf{61.63} & \textbf{58.84} & \textbf{69.77} & - \\
\hline
\end{tabular}
}
\end{table*}

\subsection{Comparison with State-of-the-art Approaches}\label{Method Performance}

\begin{table*}[ht]
\centering
\vspace{-10pt}
\caption{FSCIL performance on MiniImageNet. ``Avg imp.'' calculates our improvement over other approaches. The approaches include class-incremental or few-shot learning approaches with FSCIL setting and FSCIL approaches. }
\label{tab:mini-imagenet}
\resizebox{\linewidth}{!}{
\begin{tabular}{lccccccccccc}
\hline
	\multirow{2}{*}{Approach} & \multicolumn{9}{c}{Accuracy in each session ($\%$)  $\uparrow$} & Avg & Avg \\
    &  0 & 1 & 2 & 3 & 4 & 5 & 6 & 7 & 8 & acc. & imp.\\
	\hline
        D-Cosine \cite{vinyals2016nips} (NeurIPS'16) & 70.37 & 65.45 & 61.41 & 58.00 & 54.81 & 51.89 & 49.10 & 47.27 & 45.63 & 55.99 & +14.27\\
	iCaRl \cite{rebuffi2017cvpr} (CVPR'17) & 61.31 & 46.32 & 42.94 & 37.63 & 30.49 & 24.00 & 20.89 & 18.80 & 17.21 & 33.29 & +36.97 \\
        NCM \cite{hou2019cvpr} (CVPR'19) & 61.31 & 47.80 & 39.30 & 31.90 & 25.70 & 21.40 & 18.70 & 17.20 & 14.17 & 30.83 & +38.18 \\
        \hline
        TOPIC \cite{tao2020cvpr} (CVPR'20) & 61.31 & 50.09 & 45.17 & 41.16 & 37.48 & 35.52 & 32.19 & 29.46 & 24.42 & 39.64 & +33.62 \\
        IDLVQ \cite{chen2021iclr} (ICLR'21) & 64.77 & 59.87 & 55.93 & 52.62 & 49.88 & 47.55 & 44.83 & 43.14 & 41.84 & 51.16 & +19.10 \\
        SPPR \cite{zhu2021cvpr} (CVPR'21) & 61.45 & 63.80 & 59.53 & 55.53 & 52.50 & 49.60 & 46.69 & 43.79 & 41.92 & 52.76 & +17.50 \\
        CEC \cite{zhang2021cvpr} (CVPR'21) & 72.00 & 66.83 & 62.97 & 59.43 & 56.70 & 53.73 & 51.19 & 49.24 & 47.63 & 57.75 & +12.51 \\
        LIMIT \cite{zhou2022tpami} (TPAMI'22) & 72.32 & 68.47 & 64.30 & 60.78 & 57.95 & 55.07 & 52.70 & 50.72 & 49.19 & 59.06 & +11.2 \\
        MetaFSCIL \cite{chi2022cvpr} (CVPR'22) & 72.04 & 67.94 & 63.77 & 60.29 & 57.58 & 55.16 & 52.90 & 50.79 & 49.19 & 58.85 & +11.41 \\
        C-FSCIL \cite{hersche2022cvpr} (CVPR'22) & 76.40 & 71.14 & 66.46 & 63.29 & 60.42 & 57.46 & 54.78 & 53.11 & 51.41 & 61.61 & +8.65\\
        DF Replay \cite{liu2022eccv} (ECCV'22) & 71.84 & 67.12 & 63.21 & 59.77 & 57.01 & 53.95 & 51.55 & 49.52 & 48.21 & 58.02 & +12.24\\
        ALICE \cite{peng2022eccv} (ECCV'22) & 80.60 & 70.60 & 67.40 & 64.50 & 62.50 & 60.00 & 57.80 & 56.80 & 55.70 & 63.99 & +6.27\\
        DSN \cite{yang2023tpami} (TPAMI'23) &68.95 &63.46 &59.78 &55.64 &52.85 &51.23 &48.9 &46.78 &45.89 &54.83 &+15.45\\
        MCNet~\cite{ji2023tip} (TIP'23)&72.33&67.70 &63.50 &60.34 &57.59 &54.70 &52.13 &50.41 &49.08 &58.64 &+11.62 \\
        TEEN \cite{wang2023nips} (NeurIPS'23) &73.53 &70.55 &66.37 &63.23 &60.53 &57.95 &55.24 &53.44 &52.08 &61.44 &+8.82 \\
        NC-FSCIL \cite{yang2023iclr} (ICLR'23) & 84.02 & 76.80 & 72.00 & 67.83 & 66.35 & 64.04 & 61.46 & 59.54 & 58.31 & 67.82 & +2.44\\
        GKEAL \cite{zhuang2023cvpr} (CVPR'23) &73.59 &68.90 &65.33 &62.29 &59.39 &56.70 &54.20 &52.59 &51.31 &60.48 &+9.78 \\
        CABD \cite{zhuang2023cvpr} (CVPR'23) &74.65 &70.70 &66.81 &63.63 &61.36 &58.14 &55.59 &54.23 &53.39 &62.06 &+8.20\\
        OSHHG \cite{cui2024tmm} (TMM'24)&60.65&59.00 &56.59 &54.78 &53.02 &50.73 &48.46 &47.34 &46.75 &53.04 &+17.22\\ 
        EHS \cite{deng2024wacv} (WACV'24)	&69.43&64.86 &61.30 &58.21 	&55.49 	&52.77&50.22 &48.61 &47.67 &56.51 &+13.75 \\
        OrCo~\cite{ahmed2024cvpr} (CVPR'24) & 83.30 & 75.32 & 71.53 & 68.16 & 65.63 & 63.12 & 60.20 & 58.82 & 58.08 & 67.13  & +1.60 \\
         \textbf{Our GCMR} & \textbf{87.38}  & \textbf{77.44} & \textbf{74.45} & \textbf{71.11}  & \textbf{68.80} & \textbf{66.59} & \textbf{64.45} & \textbf{62.23} & \textbf{59.85} & \textbf{70.26} & - \\
\hline
\end{tabular}
}
\vspace{-10pt}
\end{table*}

\begin{table*}[ht]
\vspace{-10pt}
\caption{FSCIL performance on CUB200. ``Avg imp.'' calculates our improvement over other approaches. The approaches include class-incremental or few-shot learning approaches with FSCIL setting, and FSCIL approaches.}
\label{tab:cub200}
\centering
\resizebox{\linewidth}{!}{
\begin{tabular}{lccccccccccccc}
\hline
	\multirow{2}{*}{Approach} & \multicolumn{11}{c}{Accuracy in each session ($\%$)   $\uparrow$} & Avg & Avg \\
    &  0 & 1 & 2 & 3 & 4 & 5 & 6 & 7 & 8 & 9 & 10 & acc. & imp.\\
	\hline
        D-Cosine \cite{vinyals2016nips} (NeurIPS'16) & 75.52 & 70.95 & 66.46 & 61.20 & 60.86 & 56.88 & 55.40 & 53.49 & 51.94 & 50.93 & 49.31 & 59.36 & +8.25\\
        iCaRl \cite{rebuffi2017cvpr} (CVPR'17) & 68.68 & 52.65 & 48.61 & 44.16 & 36.62 & 29.52 & 27.83 & 26.26 & 24.01 & 23.89 & 21.16 & 36.67 & +30.94 \\
        NCM \cite{hou2019cvpr} (CVPR'19) & 68.68 & 57.12 & 44.21 & 28.78 & 26.71 & 25.66 & 24.62 & 21.52 & 20.12 & 20.06 & 19.87 & 32.49 & +35.12\\
       \hline
        TOPIC \cite{tao2020cvpr} (CVPR'20) & 68.68 & 62.49 & 54.81 & 49.99 & 45.25 & 41.40 & 38.35 & 35.36 & 32.22 & 28.31 & 26.28 & 43.92 & +23.69\\
        IDLVQ \cite{chen2021iclr} (ICLR'21) & 77.37 & 74.72 & 70.28 & 67.13 & 65.34 & 63.52 & 62.10 & 61.54 & 59.04 & 58.68 & 57.81 & 65.23 & +2.38\\
        SPPR \cite{zhu2021cvpr} (CVPR'21) & 68.68 & 61.85 & 57.43 & 52.68 & 50.19 & 46.88 & 44.65 & 43.07 & 40.17 & 39.63 & 37.33 & 49.32 & +18.29\\
        CEC \cite{zhang2021cvpr} (CVPR'21) & 75.85 & 71.94 & 68.50 & 63.50 & 62.43 & 58.27 & 57.73 & 55.81 & 54.83 & 53.52 & 52.28 & 61.33 & +6.28\\  
        LIMIT\cite{zhou2022tpami} (TPAMI'22) & 76.32 & 74.18 & 72.68 & 69.19 & 68.79 & 65.64 & 63.57 & 62.69 & 61.47 & 60.44 & 58.45 & 66.67 & +0.84\\
        MetaFSCIL \cite{chi2022cvpr} (CVPR'22) & 75.9 & 72.41 & 68.78 & 64.78 & 62.96 & 59.99 & 58.3 & 56.85 & 54.78 & 53.82 & 52.64 & 61.93 & +5.68\\
        FACT \cite{zhou2022cvpr} (CVPR'22) & 75.90 & 73.23 & 70.84 & 66.13 & 65.56 & 62.15 & 61.74 & 59.83 & 58.41 & 57.89 & 56.94 & 64.42 & +3.19\\
        DF replay \cite{liu2022eccv} (ECCV'22) & 75.90 & 72.14 & 68.64  &63.76 & 62.58 & 59.11 & 57.82 & 55.89 & 54.92 & 53.58 & 52.39 & 61.52 & +6.09\\
        ALICE \cite{peng2022eccv} (ECCV'22) & 77.40 & 72.70 & 70.60 & 67.20 & 65.90 & 63.40 & 62.90 & 61.90 & 60.50 & 60.60 & 60.10 & 65.75 & +1.86\\
        DSN \cite{yang2023tpami} (TPAMI'23) &76.06 &72.18 &69.57 &66.68 &64.42 &62.12 &60.16 &58.94 &56.99 &55.10 &54.21 &63.31 &4.30 \\
        MCNet~\cite{ji2023tip} (TIP'23) &77.57 &73.96 &70.47 &65.81 &66.16 &63.81 &62.09 &61.82 &60.41 &60.09 &59.08 &65.57  &+2.04 \\
        TEEN \cite{wang2023nips} (NeurIPS'23) &77.26 &76.13 &72.81 &68.16 &67.77 &64.40 &63.25 &62.29 &61.19 &60.32 &59.31 &66.63  &+0.98  \\
        NC-FSCIL \cite{yang2023iclr} (ICLR'23) & {80.45} & {75.98} & {72.30} & {70.28} & {68.17} & {65.16} & \textbf{64.43} & {63.25} & {60.66} & {60.01} & {59.44} & {67.28} & +0.33\\
        GKEAL \cite{zhuang2023cvpr} (CVPR'23) &78.88 &75.62 &72.32 &68.62 &67.23 &64.26 &62.98 &61.89 &60.20 &59.21 &58.67 &66.35  &+1.26  \\
        CABD \cite{zhao2023cvpr}(CVPR'23) &79.12 &75.37 &72.8 &69.05 &67.53 &65.12 &64.00&\textbf{63.51}&61.87&\textbf{61.47}&\textbf{60.93}&67.34 &+0.27 \\
        OSHHG \cite{cui2024tmm} (TMM'24) &63.20 &62.61 &59.83 &56.82 &55.07 &53.06 &51.56&50.05&47.50&46.82&45.87&53.85 &+13.76 \\
        MgSvF \cite{zhao2024tpami} (TPAMI'24) & 72.29 & 70.53 & 67.00 & 64.92 & 62.67 & 61.89 & 59.63 & 59.15 & 57.73 & 55.92 & 54.33 & 62.37 & +5.24 \\
        OrCo~\cite{ahmed2024cvpr} (CVPR'24) & 75.59  & 66.85  & 64.05  & 63.69  & 62.20  & 60.38  & 60.18  & 59.20  & 58.00  & 58.42  & 57.94  & 62.41  & +5.37 \\
 \textbf{Our GCMR} & \textbf{80.64} & \textbf{76.40} & \textbf{73.68} & \textbf{70.35} & \textbf{68.40} & \textbf{65.44} & {63.93} & {62.35} & \textbf{61.88} & {60.75} & {59.87} & \textbf{67.61} & +0.00 \\
\hline
\end{tabular}
\vspace{-10pt}
}
\end{table*}

We conduct the experimental comparisons with the state-of-the-art approaches on  CIRAR100, MiniImageNet, and CUB200, and report the results in \tabref{tab:cifar100}, \tabref{tab:mini-imagenet} and \tabref{tab:cub200}, respectively. We can find several meaningful observations. 

First, GCMR consistently achieves the best performance over all sessions on MiniImageNet and CIRAR100, and the best average performance on CUB200. For example, it gets a minimal accuracy improvement by $3.36\%$ in base session and $2.44\%$ in average over all sessions in MiniImageNet, $3.00\%$ in base session and $2.27\%$ in average over all sessions in CIRAR100.  

Second, our approach GCMR shows stability in incremental learning sessions. For example, on the CUB200 benchmark, in spite of only a slight accuracy improvement of $0.19\%$ in base session over the state-of-the-art NC-FSCIL, our approach still achieves the average accuracy improvement of $0.33\%$ over all sessions. 

Third, it is noted that the finetuned model in base session is very important for subsequent incremental sessions. As expected, a highly accurate base model can make the incremental models continuously perform well. For example, our base models give an accuracy of over $80\%$ on all three benchmarks and then the incremental models all deliver an accuracy of more than $60\%$ in the following seven sessions.

Fourth, as expected, the performance improvements may largely be attributed to the superior architecture of ViT. However, we need to point out the effect of two key components in our approach: 1) generative domain adaptation finetuning plays a crucial role in improving base session, as showed in Tab.~\ref{tab:effect-dmr} and Fig.~\ref{fig:effect-finetuning-pretraining}, which can provide more adaptable representations for incremental sessions, and 2) co-memory regularization plays a role in maintaining representation consistency, thereby mitigating catastrophic forgetting and achieving stable performance in incremental sessions, as showed in \tabref{tab:cifar100}, \tabref{tab:mini-imagenet} and \tabref{tab:cub200}. 

\subsection{Ablation Study}\label{Ablation}
\begin{table*}[t]
\centering
\caption{The effect of the encoder in generative domain adaptation finetuning step.}
%\resizebox{1.8\columnwidth}{!}{
\resizebox{\linewidth}{!}{
\begin{tabular}{ccccccccccccc}
\hline
\multirow{2}{*}{Dataset} & \multirow{2}{*}{Encoder (\#Parameters)} & \multirow{2}{*}{$\beta$} & \multicolumn{9}{c}{Accuracy in each session ($\%$)  $\uparrow$} & Avg \\
    & & & 0 & 1 & 2 & 3 & 4 & 5 & 6 & 7 & 8 & acc.\\
	\hline
	\multirow{6}{*}{MiniImageNet} & ViT-B (85.9M) & 0.0 & \textbf{87.87}  & \textbf{77.52} & 72.60 & 68.03 & 64.81 & 61.33 & 58.39 & 56.55 & 54.54 & 66.84 \\
        & ViT-B (85.9M) & 1.0 & 87.85  & 77.35 & 74.20 & 70.64 & 68.64 & \textbf{66.78} & \textbf{65.53} & \textbf{63.35} & \textbf{60.61} & \textbf{70.55} \\
        & ViT-B (85.9M) & 0.8 & 87.38  & 77.44 & \textbf{74.45} & \textbf{71.11}  & \textbf{68.80} & 66.59 & 64.45 & 62.23 & 59.85 & 70.26\\
        & ViT-B (85.9M) & 0.3 & 87.35  & 77.23 & 73.04 & 70.64 & 67.04 & 62.76 & 60.09 & 58.31 & 56.86 & 68.15 \\
        & ResNet34 (21.8M) & 0.3 & 84.33 & 76.65 & 72.18 & 68.93 & 66.56 & 65.01 & 62.65 & 60.14 & 58.57 & 68.32 \\
        & ResNet18 (11.2M) & 0.3 & 84.11 & 77.11 & 72.05 & 68.40 & 66.55 & 64.57 & 62.03 & 60.03 & 58.51 & 68.15\\
        \hline
\multirow{6}{*}{CIRAR100} & ViT-B (85.9M) & 0.0 & 85.40  & 77.65 & \textbf{74.96} & 70.96 & 67.50 & 63.92 & 61.39 & 58.46 & 56.11 & 68.48 \\
        & ViT-B (85.9M) & 1.0 & 85.40  & 76.60 & 74.44 & \textbf{71.44} & 68.40 & 65.23 & 63.18 & 60.64 & 58.32 & 69.29 \\
	& ViT-B (85.9M) & 0.7 & \textbf{85.52} & \textbf{77.97} & 74.50 & 71.19 & \textbf{68.69} & \textbf{66.07} & \textbf{63.59} & \textbf{61.63} & \textbf{58.84} & \textbf{69.77} \\
  & ViT-B (85.9M) & 0.3 & 85.45 & 77.45 & 74.47 & 71.13 & 67.75 & 64.65 & 62.78 & 60.14 & 58.24 & 69.12 \\
    & ResNet34 (21.8M) & 0.3 & 84.73 & 76.90 & 73.64 & 70.17 & 67.38 & 64.64 & 62.21 & 61.23 & 58.49 & 68.82 \\
    & ResNet18 (11.2M) & 0.3 & 84.25 & 76.89 & 73.94 & 69.97 & 66.60 & 63.52 & 61.73 & 59.63 & 57.68 & 68.26 \\ 
\hline
\end{tabular}
}
\label{ep}
\end{table*}

\myPara{Effect of the encoder.} We conduct an experiment to analyze the effect of the encoder by comparing the performance with ResNet and ViT encoders and report the results in \tabref{ep}. In base session, ViT encoder always gives a higher accuracy than the the ResNet encoder on MiniImageNet and CIRAR100, showing its representation superiority. In incremental sessions, a more powerful ViT encoder provides better adaptation in subsequent training. Moreover, our memory-regularization incremental learning can facilitate the model even with ResNet, where more powerful ResNet34 delivers better results. 

\myPara{Effect of memory}. \tabref{ep} shows the performance with different $\beta$, which balances the effect of two memories. When we only use the softmax classifier, it prefers higher incremental performance in early stage but suffers from rapid decay in late sessions. Among them, weight memory contains short and variable weights and allows the increment to quickly acquire novel class knowledge. The model has high variability in acquiring different memories when weight memory is dominant, especially in new memories, causing accuracy to change greatly with incremental sessions. In this case, its operation increases the probability of forgetting old knowledge. On contrary, representation memory stores class features for a long time and is more stable in all sessions, thus the accuracy of incremental sessions decreases relatively little when this memory takes the lead. In this case, two cooperative memories focus on different directions and have high potential in incremental tasks. While combining these two memories, the accuracy decline is reduced and impacted by complementary, demonstrating that the weight and representation memory complement each other. Thus the combination balances the feature extraction of new and old classes. Furthermore, when using co-memory regularization alone and ResNet12 as the backbone encoder (as shown in Tab.~\ref{tab:effect-dmr}), our approach outperforms other completing approaches, demonstrating that the proposed co-memory regularization is effective regardless of different backbones.

\begin{figure}[t]
\centering
\vspace{-10pt}
\includegraphics[width=0.6\linewidth]{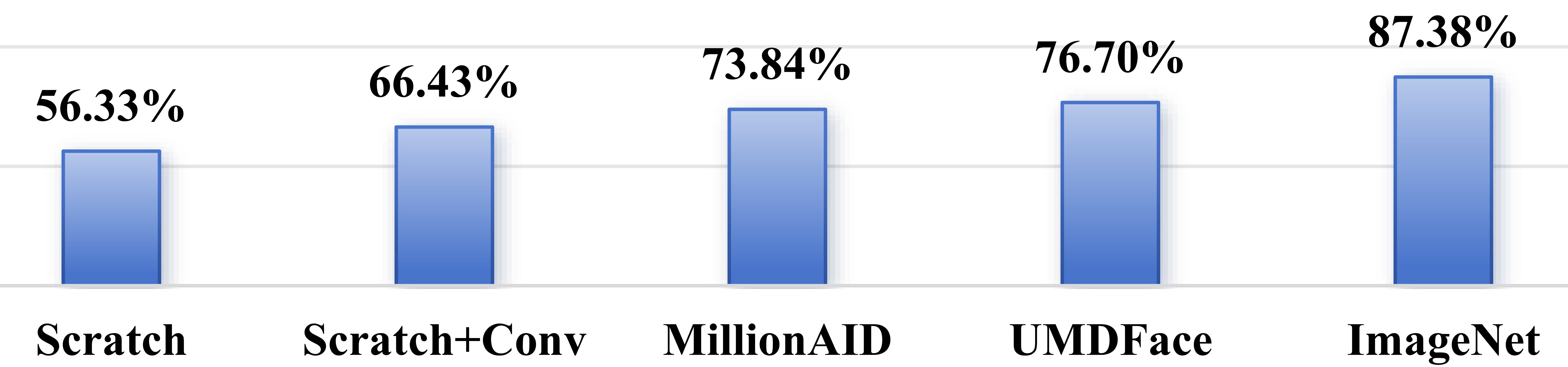}
\caption{The base accuracy on MiniImageNet with different pretraining datasets. Here, the model pretrained on ImageNet achieves the best accuracy but \emph{leaks information}.}
\label{fig:pretrain}
\vspace{-10pt}
\end{figure}

\myPara{Effect of pretraining datasets.} After the promising performance of our incremental learning method (Subsection \ref{subsec:effect-dmr}), we further study the base models pretrained on different datasets. As shown in Fig. \ref{fig:pretrain}, ViTs trained from scratch even with convolutional initialization \cite{zheng2024convolutional} (Scratch+Conv) suffer from low accuracy since such training is very data-hungry, ViTs pre-trained on specific datasets (\eg, faces~\cite{bansal2017ijcb} and remote sensing images~\cite{long2021aid}) still can not achieve satisfying accuracy, and ViT pre-trained on the dataset with similar target distribution can facilitate FSCIL. These results imply the effectiveness of large-scale pre-training for FSCIL, which is similar to \cite{brown2020nips,zhai2023iccv,clap2024cvpr} and provides a scalable solution for learning the following knowledge better.   

\begin{figure}[htbp]
\centering
\begin{minipage}[ht]{0.50\textwidth}
\centering
\vspace{-10pt}
\captionof{table}{The effect of finetuning and pretraining in base session.} 
\label{tab:effect-finetuning-pretraining} 
\resizebox{1.1\linewidth}{!}{
\begin{tabular}{ccccc}
\hline
\multirow{2}{*}{Dataset} & \multirow{2}{*}{Finetuning}  & \multirow{2}{*}{Pretraining} & Accuracy($\%$)& Accuracy($\%$)   \\ 
 & & &  (10 epoches)&(Final)\\
    \hline
    \multirow{4}{*}{MiniImageNet}  & - & - & 18.92 & 36.43\\
     &  \checkmark  & - & 20.10 & 37.40\\
     & - & \checkmark & 75.50 & 86.51\\
     & \checkmark & \checkmark & 76.10 & 87.35\\ 
    \hline
    \multirow{4}{*}{CIRAR100} & - & - & 22.48 & 41.02\\
     & \checkmark & - & 23.95 & 41.98\\ 
     &  - & \checkmark & 35.73 & 84.63\\
     &  \checkmark & \checkmark & 37.22 & 85.52\\  
    \hline
     \multirow{4}{*}{CUB200}  & - & - & 9.71 & 38.91\\
     & \checkmark & - & 10.19 & 39.51\\
     & - & \checkmark & 44.31 & 79.65\\ 
     & \checkmark & \checkmark & 45.60 & 80.64\\
\hline
\end{tabular}
\vspace{-10pt}
}
\end{minipage}
\hfill
\begin{minipage}[ht]{0.40\textwidth}
    \centering
    \vspace{5pt}
    \includegraphics[width=0.9\linewidth]{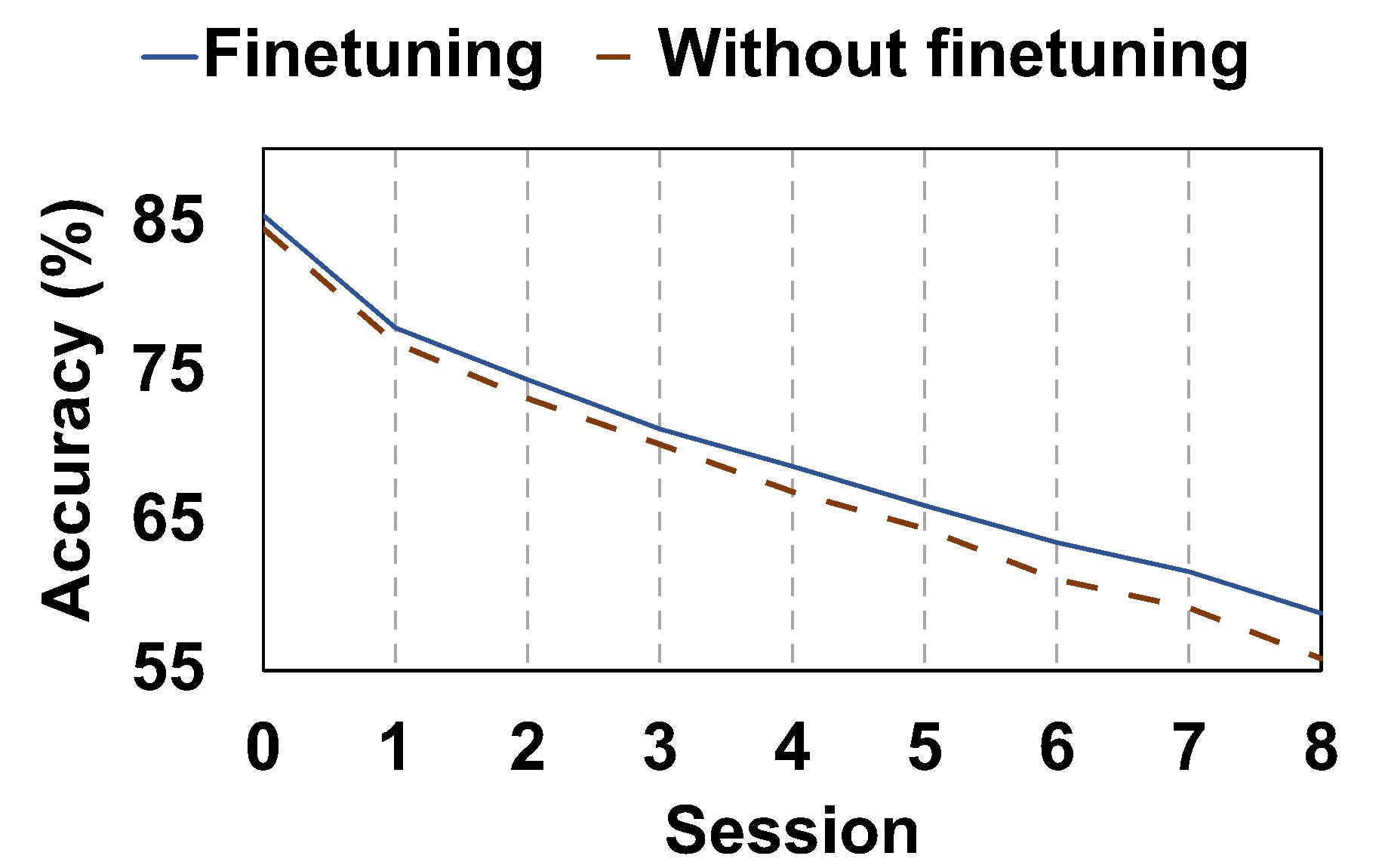}
    \captionof{figure}{The effect of finetuning in incremental sessions on CIFAR100.} 
    \vspace{-10pt}
    \label{fig:effect-finetuning-pretraining} 
  \end{minipage}
\end{figure}
\myPara{Effect of generative domain adaptation finetuning.}
As shown in \tabref{tab:effect-finetuning-pretraining}, generative domain adaptation finetuning (GDAF) leads to an accuracy improvement of $0.47$-$1.49\%$ in base session. Especially, the accuracy improvement is largely attributed to pretraining (\eg, over 40\% than that without pretraining) and further boosted by finetuning (\eg, about 1\%). It is noted that the main objective of finetuning is domain adaptation, which can mitigate the accuracy drop in incremental sessions more effectively, as shown in Fig.~\ref{fig:effect-finetuning-pretraining}.
The reasons include: 1) GDAF can guide cross-domain knowledge transfer in model parameters due to its strong ability to memorize domain features, and 2) GDAF can improve model adaptability by combining supervised classification and self-supervised reconstruction. 

\begin{figure}[ht]
\centering
\vspace{-10pt}
\includegraphics[width=1.0\linewidth]{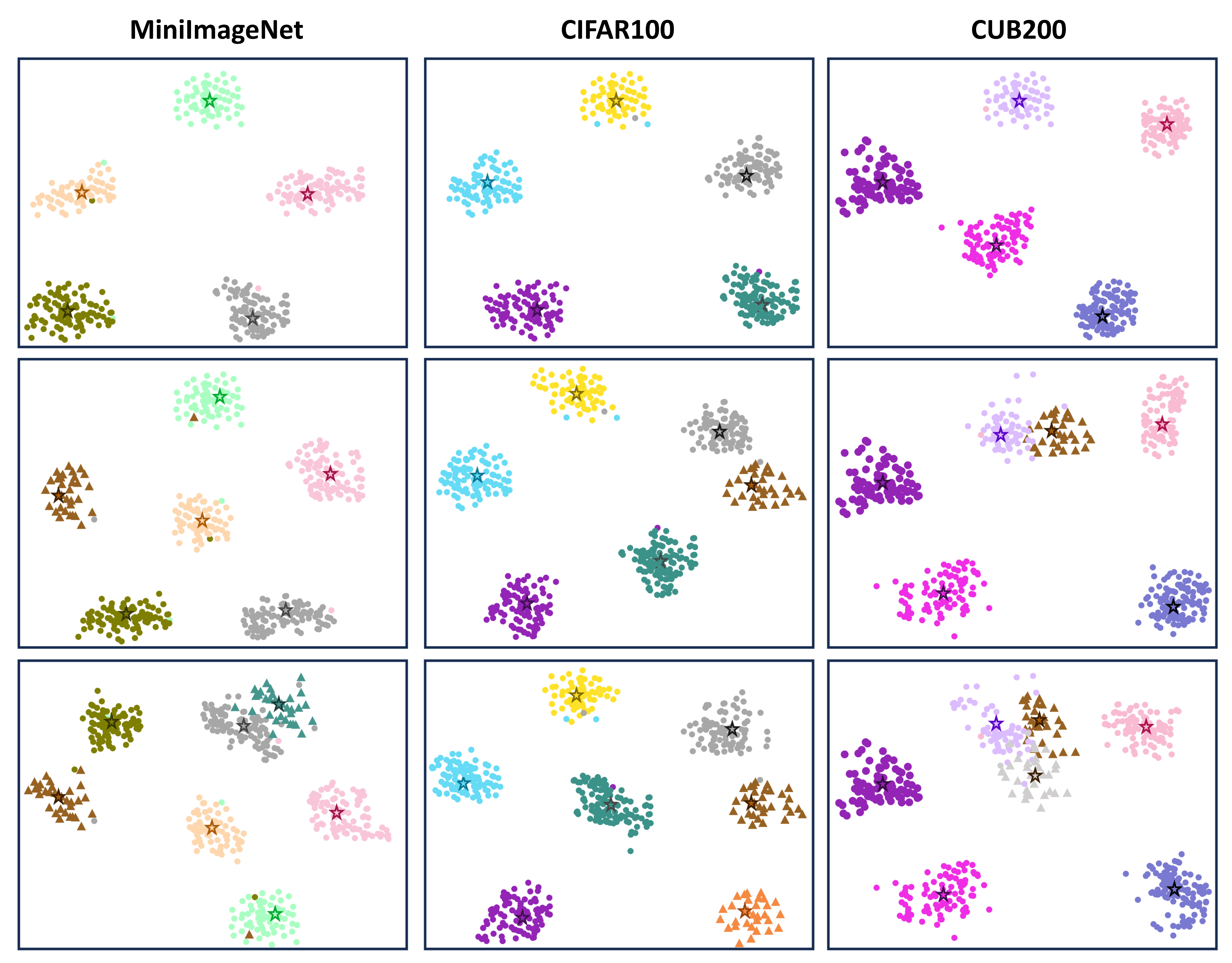}
\caption{t-SNE representation visualization. We randomly select the examples over several base and incremental classes to show how these classes change at base (top), intermediate incremental (middle) and last incremental (bottom) sessions. 
`$\bullet$', `$\blacktriangle$' and `$\bigstar$' represent base classes, incremental classes and class mean features, respectively.}
\vspace{-10pt}
\label{fig:tsne}
\end{figure}

\subsection{Further Discussion}\label{Discussion}

\myPara{Representation visualization.}~We visualize the representations in the base and incremental sessions with t-SNE~\cite{maaten2008jmlr} in \figref{fig:tsne}. We can find that the examples in the same classes can be clustered well in base session by finetuning the ViT encoder. The generalization of the ViT encoder also benefits the learning in subsequent sessions, thus the representations also show good discriminability but get more complex along with incremental learning, as expected.
Like other methods~\cite{snell2017nips,yang2023iclr,wang2024ijcv}, we also use class mean features as representation memory since they are effective and efficient to retain class knowledge. Continuous averaging may weaken previously learned class representations, particularly when novel classes are introduced in incremental sessions. However, the features still tend to reside near the center of class distribution, as shown in \figref{fig:tsne}. This property helps maintain a representative and stable feature for each old class to alleviate catastrophic forgetting. In our approach, we further train the model with memory regularization in incremental sessions to preserve feature norms, which well balances the influence between old and new class features. Our approach forces the mapping position of the examples to tend to be constant or have only a small change. However, there still exist significant changes in some class examples and class means. From the last two rows in \figref{fig:tsne}, the feature maps of some old classes are shifted, which frees up more space for novel classes. Meanwhile, our constraint on class means can maintain the clustering of example maps.

\myPara{Catastrophic forgetting.}~One key to addressing this issue is making the model retain memory like humans. Our approach freezes encoder parameters and uses memory regularization during incremental learning, which preserves the old knowledge effectively. To incrementally learn parameters for recognizing novel classes, our approach uses two cooperative memories, including representation memory with greater invariance and weight memory with greater variability. These two types of memories damage old or new knowledge but also complement each other. Therefore, as shown in \tabref{tab:mini-imagenet}, \tabref{tab:cifar100} and \tabref{tab:cub200}, the model can incrementally learn new knowledge from the stream of novel classes while preserving old knowledge.

\myPara{Overfitting.}~In all incremental sessions, it is easy to overfit to novel classes due to the examples of novel classes are not sufficient. To address that, besides calculating memories to reduce overfitting, our approach adds the ``dropout'' layer and ReLU activation function in training and also uses memory regularization to weaken the model complexity. On the one hand, our approach increased the training difficulty in incremental sessions. On the other hand, our approach can use fewer training rounds in both base and incremental sessions to learn models without overfitting.

\begin{figure}[ht]
\centering
\vspace{-10pt}
\includegraphics[width=0.6\linewidth]{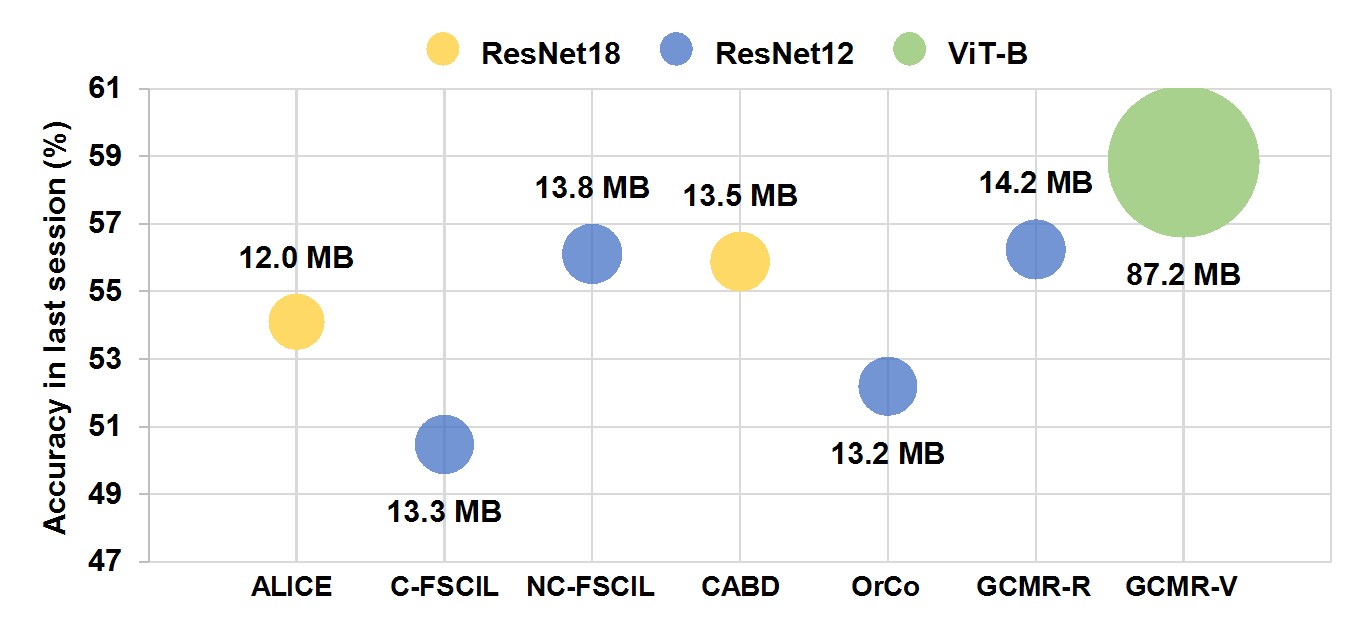}
\caption{The network architecture, memory budget and accuracy on CIFAR100. A larger area means a more memory overhead. GCMR-R and GCMR-V denote our approach with ResNet12 backbone and ViT-B backbone, respectively.}
\label{fig:structure}
\vspace{-15pt}
\end{figure}

\myPara{Resource efficiency.} Fig.~\ref{fig:structure} compares the resource overhead of our GCMR with five competing approaches, where four approaches (ALICE, C-FSCIL, NC-FSCIL and OrCo) store class mean embeddings in memory, CABD stores the network parameters in previous session, and our approach stores class mean embeddings and an extra weight memory. Our GCMR-R achieves better accuracy when costing a similar memory, and GCMR-V remarkably improves the accuracy when costing a large memory of 87.2 MB. Moreover, due to extra weight memory, the memory of our approach increases gradually with the number of learned tasks, but still remains minimal, \eg, GCMR-R takes only $0.4$ MB more than NC-FSCIL on CIFAR100 in last session.

\myPara{Limitation.}~From \tabref{tab:cub200}, it is found that our GCMR has a slightly lower accuracy than other approaches in four sessions on CUB200. We analyze that the old and novel classes are highly similar, which leads to a close distance in the memory space and a challenge in accurately distinguishing objects during ViT finetuning. The high similarity of some classes leads to the excessive overlap of the feature mapping between the old and the new examples, confusing the old and new knowledge. Therefore, we may add new distance constraints between old and novel classes, or pre-allocate space to avoid class confusion in the future. Moreover, \figref{fig:pretrain} shows that the base accuracy is very affected by the pretraining dataset, implying that large domain discrepancy between pretraining and training data will limit model learning. Furthermore, although ViT backbone contributes to the overall performance and provides a stronger feature representation capability than ResNet, it takes more model parameters and usually needs pretraining with self-supervised MAE under limited base examples.

\begin{table}[ht]
\centering
\vspace{-10pt}
\caption{Performance under large model pretraining.}
\begin{tabular}{ccccccc}
\hline
\multirow{2}{*}{Approach} & \multicolumn{3}{c}{Accuracy in the base / last session (\%)}   \\ 
  &CIFAR100 & MiniImageNet&CUB200\\
    \hline
  Our GCMR & 85.52 / 58.84 & 87.38 / 59.85 & 80.64 / 59.87\\
  LIMIT+SV~\cite{qiu2023icme} &86.77 / 69.75&90.55 / 81.65&84.19 / 76.17\\
  CPE-CLIP~\cite{alessandro2023iccvw} & 87.83 / 80.52 &90.23 / 82.77 & 81.58 / 64.60 \\
  PriViLege~\cite{park2024cvpr}  &90.88 / 86.06& 96.68 / 94.10&82.21 / 75.08\\
\hline
\end{tabular}
\vspace{-10pt}
\label{dt}
\end{table}

\myPara{Pretraining with large models.} Recently, several works employed large pretrained models to facilitates FSCIL task. As shown in Tab.~\ref{dt}, the approaches with large pretrained vision and language Transformers (LIMIT+SV~\cite{qiu2023icme}, CPE-CLIP~\cite{alessandro2023iccvw} and PriViLege~\cite{park2024cvpr}) deliver remarkable performance on all three benchmarks, where the learned representations provide high accuracy in base session and low accuracy drop in incremental sessions. We suspect that large models can learn more essential and generalizable features by incorporating 
sufficient knowledge (\eg, language or prompt). Thus, our further work will follow this tendency and learn or select essential representations on few data to mitigate catastrophic forgetting in FSCIL task.

\section{Conclusion}
This paper proposes a generative co-memory regularization approach to facilitate few-shot class-incremental learning. The approach first takes generative domain adaptation finetuning to finetune a pretrained encoder by jointly optimizing self-supervised masked feature reconstruction and supervised feature classification. Then, it performs memory-regularized incremental learning to incrementally train a classifier head by simultaneously feature classification and feature regularization assisted by two cooperative memories. The two memories are created and updated during training, collaboratively regularizing the incremental learning, which effectively mitigates catastrophic forgetting and overfitting. Extensive experiments clearly show the effectiveness of our approach. Our future works include more effective methods in domain adaptation and more applications in practical scenarios like remote sensing object understanding and medical tumour recognition.  

\myPara{Acknowledgements}
This work was partially supported by grants from the Pioneer R\&D Program of Zhejiang Province (2024C01024) and Open Research Project of National Key Laboratory of Science and Technology on Space-Born Intelligent Information Processing (TJ-02-22-01).

\section*{Declarations}
%\bmhead{Funding} Pioneer R\&D Program of Zhejiang Province (2024C01024), and Open Research Project of National Key Laboratory of Science and Technology on Space-Born Intelligent Information Processing (TJ-02-22-01).

\bmhead{Competing interests}
The authors have no competing interests to declare that are relevant to the content of this article. 

\bmhead{Data Availability}
All the data used in this paper are publicly available, which contains MiniImageNet \cite{deng2009cvpr}, CIRAR100~\cite{krizhevsky2019cifar} and CUB200~\cite{Wah2011TheCB}. 

%\bmhead{Author contribution}
% Material preparation, experiments and analysis were performed by Kexin Bao and Shiming Ge. Yong Li and Dan Zeng participated in the discussion and analysis. The first draft of the manuscript was written by Shiming Ge and all authors commented on previous versions of the manuscript. All authors read and approved the final manuscript.

%\bmhead{Materials availability}
% Not applicable.

%\begin{appendices}

%\end{appendices}

%\bibliographystyle{sn-mathphys-num.bst}
\bibliography{ijcv}% common bib file

\end{document}